\documentclass[letterpaper, 10 pt, conference]{ieeeconf}  
\usepackage{graphicx}%用于插入和管理图像,包括缩放、旋转、裁剪等功能
\usepackage{amsmath}%提供增强的数学公式功能,包括多行公式、分段函数、对齐环境等
\usepackage{amssymb}%提供额外的数学符号,如 R、Z 等黑板粗体字母
\usepackage{amsfonts}%提供额外的数学字体,如 a（花体字母N（黑板粗体）
\usepackage{flushend} %确保双栏文档的最后一页内容对齐
\usepackage{tikz}
\usepackage{caption}
\usepackage{adjustbox} % 用于旋转文本
\usepackage{lipsum} % 用于生成示例文本
\usepackage{cite}
\usepackage[colorlinks=true, urlcolor=blue]{hyperref}
\usepackage{multirow} % Required for multirows
\IEEEoverridecommandlockouts                             
\title{\LARGE \bf
GARAD-SLAM: 3D GAussian splatting for Real-time Anti Dynamic SLAM
}

\author{$^{\dagger}$Mingrui Li$^{1}$, $^{\dagger}$Weijian Chen$^{2}$, Na Cheng$^{1}$, Jingyuan Xu$^{1}$, Dong Li$^{3}$$^{*}$ and Hongyu Wang$^{1}$$^{*}$
\thanks{Corresponding authors: Dong Li and Hongyu Wang.}% <-this % stops a space
\thanks{\raggedright$^{\dagger}$Mingrui Li and $^{\dagger}$Weijian Chen contributed equally. }%
\thanks{$^{1}$Mingrui Li, Na Cheng, Jingyuan Xu, Hongyu Wang are with the School of Information and Communication Engineering, Dalian University of Technology, Dalian, 116024, China
        {\tt\small 2905450254@mail.dlut.edu.cn} }%
\thanks{$^{2}$Weijian Chen is with the School of Aeronautics and Astronautics, Sun Yat-sen University, Shenzhen, 528406, China.
        }%
\thanks{$^{3}$Dong Li is with the Faculty of Science and Technology, University of Macau, Macao, 999078, China.
        }%
}

\begin{document}

\maketitle
\thispagestyle{empty}
\pagestyle{empty}

%%%%%%%%%%%%%%%%%%%%%%%%%%%%%%%%%%%%%%%%%%%%%%%%%%%%%%%%%%%%%%%%%%%%%%%%%%%%%%%%
\begin{abstract}

The 3D Gaussian Splatting (3DGS)-based SLAM system has garnered widespread attention due to its excellent performance in real-time high-fidelity rendering. However, in real-world environments with dynamic objects, existing 3DGS-based SLAM systems often face mapping errors and tracking drift issues. To address these problems, we propose GARAD-SLAM, a real-time 3DGS-based SLAM system tailored for dynamic scenes. In terms of tracking, unlike traditional methods, we directly perform dynamic segmentation on Gaussians and map them back to the front-end to obtain dynamic point labels through a Gaussian pyramid network, achieving precise dynamic removal and robust tracking. For mapping, we impose rendering penalties on dynamically labeled Gaussians, which are updated through the network, to avoid irreversible erroneous removal caused by simple pruning. Our results on real-world datasets demonstrate that our method is competitive in tracking compared to baseline methods, generating fewer artifacts and higher-quality reconstructions in rendering.
\end{abstract}

%%%%%%%%%%%%%%%%%%%%%%%%%%%%%%%%%%%%%%%%%%%%%%%%%%%%%%%%%%%%%%%%%%%%%%%%%%%%%%%%
\vspace{-2pt}  % 调整为合适的负值,减少下方间距
\section{INTRODUCTION}
\vspace{-1pt}  % 调整为合适的负值,减少下方间距
Currently, the outstanding performance of 3DGS in photorealistic high-fidelity reconstruction has attracted significant attention, leading to the emergence of 3DGS-based SLAM systems \cite{1,2,3,deng2024compact,huang2024photo}. For fields like AR/VR and robotics, a SLAM system with real-time high-fidelity dense reconstruction capabilities holds great significance. On the other hand, existing 3DGS-based SLAM systems such as SplaTAM \cite{4} and MonoGS \cite{2} exhibit limitations when confronted with real-world environments. Real-world environments often contain a large number of moving objects, and traditional SLAM systems usually employ semantic-based methods to remove these interferences \cite{5,6,he2023ovd,liu2021rds,hu2022cfp,du2020accurate,huang2020clustervo,dai2020rgb}. However, current 3DGS-based SLAM systems cannot simply apply these feature-based removal or optical flow segmentation methods. These methods can only correct tracking errors but cannot resolve artifacts caused by dynamic noise inputs.

\begin{figure}[htbp]
    \centering
\begin{minipage}[t]{0\textwidth}
        \centering
        \raisebox{2.3cm}[0pt][0pt]{%
            \adjustbox{valign=t}{\rotatebox{90}{\makebox[2cm][c]{\fontsize{9}{11}\selectfont\textbf{ Input }}}}
        }\\
        \raisebox{-0.4cm}[0pt][0pt]{%
            \adjustbox{valign=t}{\rotatebox{90}{\makebox[2cm][c]{\fontsize{9}{11}\selectfont\textbf{ Reconstruction}}}}
        }
    \end{minipage}%
    \begin{minipage}[t]{0.50\textwidth}
        \centering
        \renewcommand{\tabcolsep}{01pt} % 减小列间距
        \begin{tabular}{cccccc}
        \includegraphics[width=4cm,height=3cm]{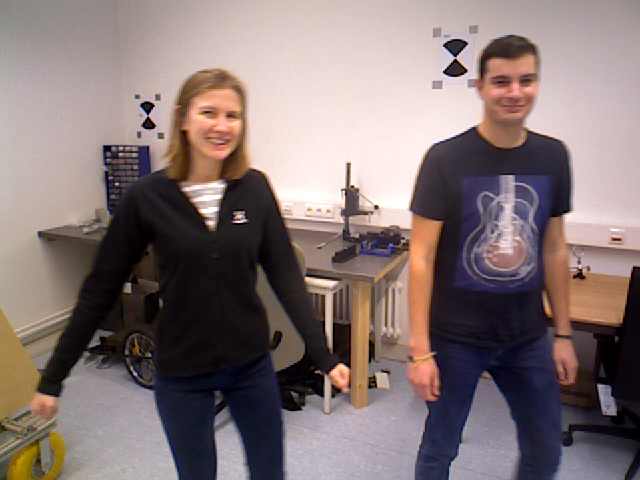}&
        \includegraphics[width=4cm,height=3cm]{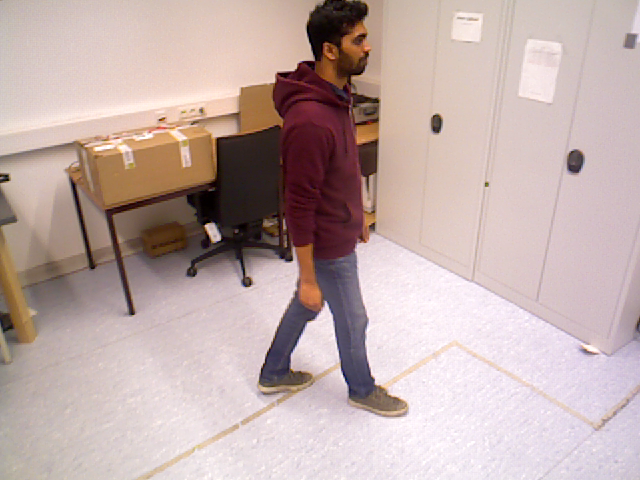}\\
        \includegraphics[width=4cm,height=3cm]{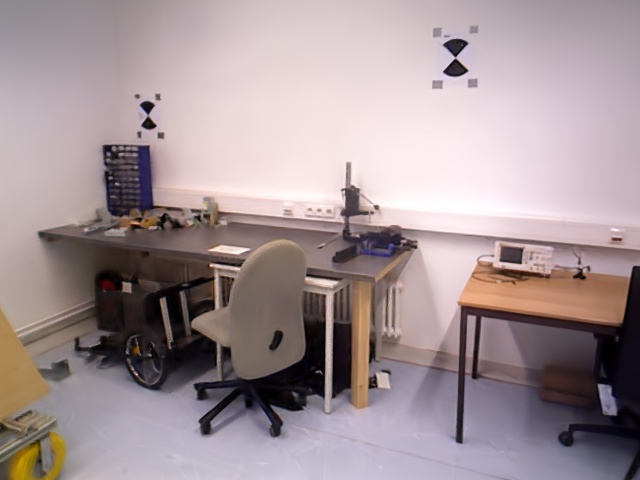}&
        \includegraphics[width=4cm,height=3cm]{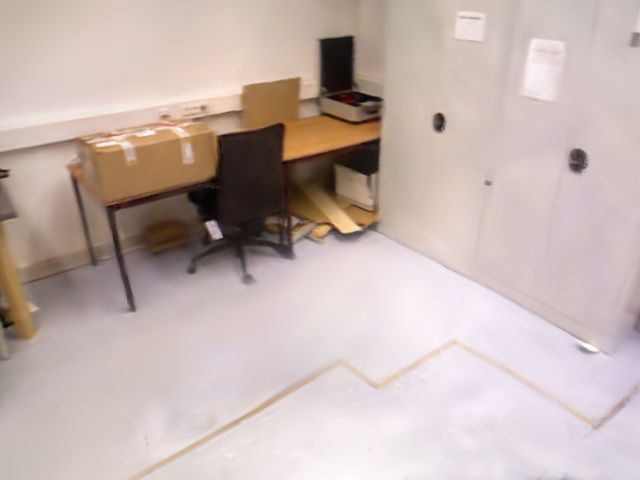}\\
        \end{tabular}
    \end{minipage}
    \caption{Rendering result in BONN datasets. The reconstruction result is generated by our GARAD-SLAM.}
    \vspace{-15pt}  % 调整为合适的负值,减少下方间距
    \label{fig1}
\end{figure}

Recently, SLAM systems based on neural implicit representations have made progress in addressing these dynamic interferences, such as DN-SLAM \cite{7}, NID-SLAM \cite{8}, DDN-SLAM \cite{DDN}, and RoDyn-SLAM \cite{10}. However, neural implicit methods still have some unresolved challenges. For example, DN-SLAM and DDN-SLAM rely on traditional front-end pose correction, and while DDN-SLAM adds additional rendering penalties during the back-end rendering process, it fails to resolve artifacts from multi-view synthesis fundamentally. NID-SLAM and RoDyn-SLAM use optical flow-based estimation methods for tracking, where RoDyn-SLAM removes sampled rays with dynamic labels while applying warp loss to track camera motion. However, these methods have limitations in real-time performance and scene geometry accuracy, with tracking and mapping remain loosely coupled.

Therefore, we propose GARAD-SLAM, a 3DGS-based SLAM system designed for dynamic environments. Unlike traditional methods, we do not perform semantic segmentation on the feature points in the tracking phase. Instead, we directly build a conditional random field (CRF) \cite{11} based on feature and position for Gaussian rendering in the back-end and perform segmentation. Using our Gaussian pyramid network, we map this segmentation back to the front-end feature points and apply labels. This allows us to remove back-end rendering artifacts while correcting pose errors. To ensure the accuracy of dynamic Gaussian labels, we build a sparse optical flow validation from the front-end feature points to recover incorrectly labeled Gaussians in the back-end. This results in tight integration of tracking and rendering in our system. Finally, we gradually reconstruct the static scene through rendering loss, eliminating the influence of dynamic Gaussians. Our test results on several real-world datasets demonstrate that our method achieves state-of-the-art results in rendering accuracy and runtime performance.

Our contributions can be summarized as follows:
\begin{enumerate}
    \item We introduce GARAD-SLAM, which to the best of our knowledge, is the first 3DGS-based SLAM system designed specifically for dynamic scenes.
    \item We achieve tight coupling and mutual enhancement between tracking and mapping, in contrast to traditional methods that only focus on tracking correction.
    \item We propose a dynamic point removal method based on a conditional CRF for Gaussians, and introduce a rendering loss for dynamic Gaussian labels. Additionally, we provide an error label recovery method that utilizes sparse optical flow for feature points corresponding to the Gaussians.
    \item Our experimental results on real-world datasets demonstrate that our method achieves state-of-the-art performance in terms of tracking and rendering accuracy.
\end{enumerate}

\section{RELATED WORK}

Neural Radiance Fields (NeRF) \cite{12} and 3D Gaussian Splatting (3DGS) \cite{13} have garnered widespread attention in recent years as effective methods for scene reconstruction \cite{deng2024plgslam,deng2024neslam,zhu2022nice,johari2023eslam,wang2023co}. These approaches have shown exceptional performance, particularly in real-time photorealistic rendering, and their integration with RGB-D SLAM systems has demonstrated impressive results. iMAP \cite{14} was the first SLAM system based on NeRF, but its single MLP model limited its ability to represent complex scenes. However, with the emergence of 3DGS-based SLAM, neural implicit representation faced significant challenges as 3DGS-based SLAM achieved better results in both tracking accuracy and rendering. 3DGS strikes a balance between implicit and explicit representations. For instance, SplaTAM \cite{4} uses frame-to-frame tracking and per-pixel output to represent 3DGS scenes, while MonoGS \cite{2} enables pure RGB scene tracking and reconstruction. Nonetheless, these SLAM systems struggle in dynamic real-world scenarios, as dynamic interferences lead to erroneous feature matching or pixel misalignment, resulting in tracking failures or loss. Furthermore, dynamic noise in the input causes rendering artifacts during mapping, posing a significant challenge to current SLAM systems.

As NeRF-SLAM continues to evolve, researchers are increasingly focusing on handling dynamic scenes. DN-SLAM \cite{7} employs optical flow estimation to remove dynamic feature points from the front-end, correcting tracking errors similarly to traditional methods. However, it overlooks the rendering artifacts caused by noisy inputs in the back end. DDN-SLAM \cite{DDN} combines deep learning-based detection with depth segmentation and introduces additional rendering penalties. RoDyn-SLAM \cite{10} uses optical flow estimation on keyframes to obtain dynamic masks, removing dynamic rays for updates, but its real-time performance and rendering accuracy are limited. NID-SLAM \cite{8} also utilizes optical flow estimation to generate dynamic masks and background completion to mitigate dynamic noise interference, but its tracking performance is constrained. Our approach integrates the strengths of these methods while addressing the issue of decoupled tracking and mapping. We use a Gaussian pyramid network to apply front-end feature points to Gaussians for densification and directly apply dynamic segmentation for rendering correction on Gaussians, mapping them back to the tracked feature points. This method tightly integrates the system, allowing tracking and mapping to mutually enhance each other, ultimately constructing a static scene map.

\begin{figure*}[t]
  \centering
  \includegraphics[width=\hsize]{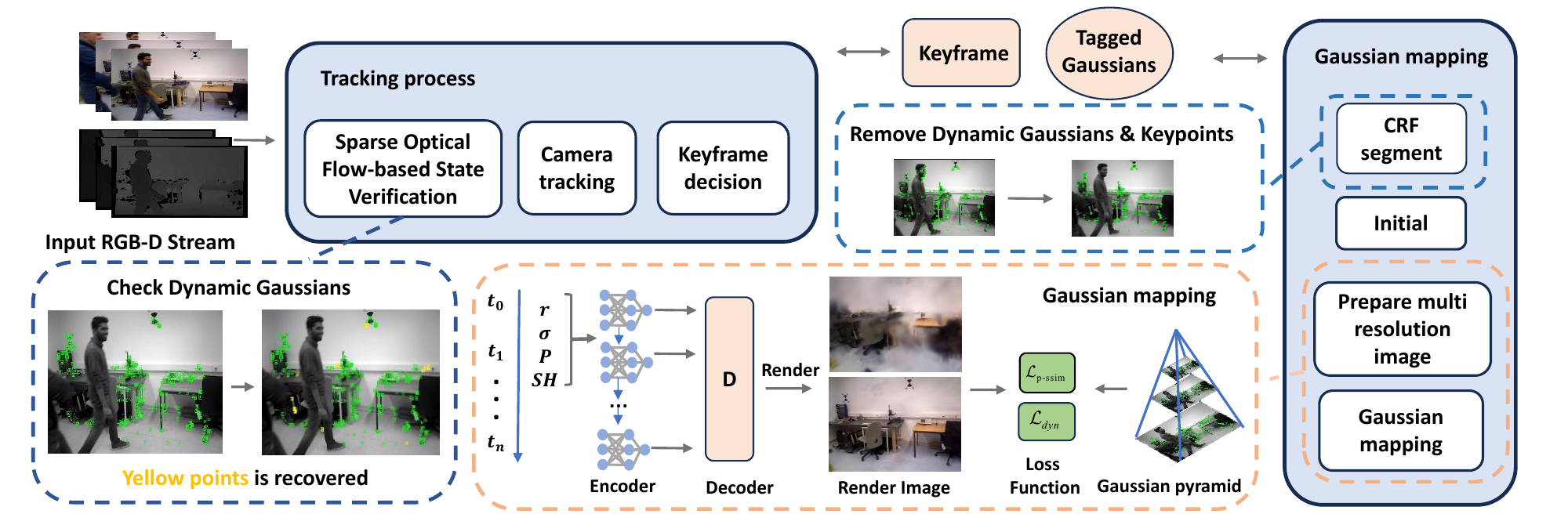}
  \caption{Frame of GARAD-SLAM. Given a series of RGB-D frames, we simultaneously construct the gaussian map and camera pose via Gaussian pyramid with Photometric-SSIM Loss $\lambda _{p-ssim}$ and dyn Loss $ \lambda _{dyn}$.  }
  \vspace{-15pt}  % 调整这个数值来减少空白
  \label{fig2}
\end{figure*}

\section{METHOD}

The framework of our system is shown in Fig. \ref{fig2}. In tracking and gaussian mapping threads, we maintain Tagged Gaussians, which present a point cloud $P_i$ that contains ORB-matched feature points $p_i$, rotation $r$, density $\sigma $, and spherical harmonics descriptors $SH$. These Tagged Gaussians allow the system to optimize tracking using graph-based optimization, while simultaneously learning the corresponding mapping by backpropagating the loss between the original images and rendering images. In Sec \ref{sec:1}, we proposed a method for determining dynamic labels of Gaussians using CRF, while also employing a Gaussian Pyramid Network to back-project and filter feature points on the front-end. In Sec \ref{sec:2}, we developed a sparse optical flow detection to further filter feature points and recover incorrect Gaussian labels. In Sec \ref{sec:3}, we introduced dynamic rendering penalties, utilizing updates from the Gaussian Pyramid Network to progressively optimize Gaussians with dynamic labels, ultimately achieving static rendering results.

\subsection{Dynamic Gaussian Validation Based on CRF} \label{sec:1}
In traditional semantic SLAM methods, dynamic objects are often handled by removing feature points associated with them in the front-end \cite{hu2022cfp,du2020accurate,huang2020clustervo,yin2022dynam,fan2022blitz,li2021dp}. This helps correct the erroneous feature matches caused by dynamic objects. Subsequently, dynamic objects are removed through pixel-level segmentation, ensuring accurate back-end mapping. However, in SLAM systems based on 3DGS, while pixel-level segmentation can effectively remove dynamic interference, it often incurs significant computational costs and may inadvertently eliminate some static Gaussians, thereby affecting the quality of back-end mapping. Conversely, the absence of pixel-level segmentation may result in residual transient interference from dynamic objects, ultimately leading to rendering artifacts.

To address this issue, we propose a method that integrates a Gaussian pyramid network with CRF segmentation. A Gaussian pyramid is a multi-scale representation generated by applying Gaussian smoothing and downsampling to the original image, capturing varying levels of detail. Unlike traditional approaches, we directly construct a CRF based on the Gaussians in the back-end and eliminate dynamic Gaussians during the pyramid optimization process. Once dynamic Gaussians are identified, we generate a Gaussian mask and map it back to the front-end feature points through the Gaussian pyramid network, correcting pose errors in front-end tracking. This method effectively removes artifact interference from the rendering process.

In particular, the system assumes a static scene for tracking during the first ten frames, generating Tagged Gaussians. Since the accumulated error is minimal in these initial frames, the pose is accurate enough for initialization. In the Gaussian mapping thread, we construct a CRF on the Tagged Gaussians, with a fully connected graph \cite{11} linking each pair of Gaussians, where $x_i = L_i \in {0, 1}$ (0 for static, 1 for dynamic). The CRF optimization process aims to minimize the Gibbs energy function:
\vspace{-2pt}  % 调整为合适的负值,减少下方间距
\begin{equation}
  \label{eq1}
E\left( X \right) =\sum_i{\psi _u\left( x_i \right)}+\sum_{i<j}{\psi _p(x_i,x_j)}
\end{equation}
Where $\psi _u\left( x_i \right)$ is the unary potential and $\psi _p(x_i,x_j)$ is the pairwise potentials.
\vspace{-2pt}  % 调整为合适的负值,减少下方间距
Based on the characteristics of Dynamic and Static Gaussians, we make the following assumptions: (1) Dynamic Gaussians typically have larger reprojection errors; (2) they show greater depth variation due to their movement; (3) Because of the RANSAC algorithm, they are observed fewer times; and (4) their epipolar distances are larger as they change positions between frames.

These assumptions can be represented by Gaussian models $\mathcal{N} _{i}^{\alpha}$, $\mathcal{N} _{i}^{\beta}$, $\mathcal{N} _{i}^{\gamma}$, $\mathcal{N} _{i}^{\delta}$, corresponding to static reprojection, depth, observation count, and epipolar distance. These are then combined into a Gaussian Mixture Model (GMM):

\begin{equation}
  \label{eq2}
P_{i}^{static}=\sum_{k=1}^4{\pi _k}\mathcal{N} (x|\mu _k,\sigma _{k}^{2})
\end{equation}
Where, $\pi _k$ denotes the weight coefficient of the $k-th$ Gaussian distribution and $\sum_{k=1}^4{\pi _k=1}$, $\mathcal{N} (x|\mu _k,\sigma _{k}^{2})$ is the $k-th$ Gaussian distribution, with $\mu _k$ as the mean and $\sigma _{k}^{2}$ as the variance.

Then the unary potential function is defined as:

\begin{equation}
  \label{eq3}
\psi _u\left( x_i \right) = -\log \left( P_{i}^{static} \right) 
\end{equation}

The pairwise potential function can be described as:
\begin{equation}
  \label{eq4}
\psi _p(x_i,x_j)=\mu (x_i,x_j)\sum_m{\omega ^mk^m(f_i,f_j)}
\end{equation}
Where $\mu(x_i,x_j)$ is a simple Potts model. $k^m(f_i,f_j)$ represents the similarity between neighboring Tagged Gaussians, $f_i,f_j$ are the feature vectors of Tagged Gaussians $i$, $j$.

% For the pairwise potential function, we use two Gaussian kernel functions to describe the feature and positional relationships between neighboring Tagged Gaussians. Static Gaussians are observed more frequently across multiple keyframes and exhibit smaller average reprojection errors. The feature kernel is defined as:

\begin{equation}
  \label{eq5}
k^1(f_i,f_j)=\exp\mathrm{(}-\frac{|\alpha _i-\alpha _j|^2}{2\sigma _{\alpha}^{2}}-\frac{|\gamma _i-\gamma _j|^2}{2\sigma _{\gamma}^{2}})
\end{equation}
Where $\alpha _i$, $\alpha _j$, represent the average reprojection errors of Tagged Gaussians $i$, $j$, while $\gamma _i$, $\gamma _j$ denote the number of observations.

The second kernel function is the position kernel, which is used to describe the spatial relationships between neighboring Tagged Gaussians. It is defined as follows:

\begin{equation}
  \label{eq6}
k^2(f_i,f_j)=\exp\mathrm{(}-\frac{|P_{i}^{p}-P_{j}^{p}|}{2\sigma _{P}^{2}}-\frac{|p_i-p_j|}{2\sigma _{p}^{2}})
\end{equation}
Where $P_{i}^{p},P_{j}^{p}$ represent the positions of Tagged Gaussians $i$ and $j$. $p_i,p_j$ denotes the corresponding pixel coordinates.

After the label assignment, we optimize only Static Gaussians. For Dynamic Gaussians, we use a time-window retention strategy to temporarily keep detected dynamic Gaussians, avoiding premature deletion of potential static ones. This delay allows for further verification in subsequent frames. The process checks label changes within a fixed frame window $n$, and Gaussians are deleted based on:
\begin{equation}
  \label{eq7}
S_i(i+n)=1-\frac{1}{n}\sum_{k=i}^{i+n}{1-L_i\left( k \right)}_i
\end{equation}
where $S_i(i+n) = 0$ represents removal of the Gaussian $i$.

\subsection{Sparse Optical Flow-based Gaussian Motion State Verification}  \label{sec:2}

Although most dynamic Gaussians can be effectively verified through CRF, there may still be instances of labeling errors. For example, dynamic Gaussians moving along the epipolar line might be missed, or some static Gaussians near the edges could be mistakenly classified as dynamic, resulting in incorrect labels. In back-end rendering, the erroneous deletion or retention of dynamic Gaussians can compromise the integrity of scene reconstruction or create visual artifacts.

To address this issue, we propose a sparse optical flow-based motion state verification method for dynamic Gaussians. This method maps dynamic Gaussians to the front-end's dynamic points through masks and uses sparse optical flow verification to accurately identify and recover those Gaussians that were incorrectly labeled as dynamic. Sparse optical flow verification reduces computational complexity while fully utilizing optical flow information between keyframes, ensuring that mistakenly deleted static Gaussians can be effectively recovered.

Specifically, we process the feature points in the front-end that correspond to the Tagged Gaussians. Through the 2D-2D matching relationship, we obtain matching point sets $P^{\mathrm{i}}=\big\{ p_{k}^{i}|k=1,2,...,n \big\} $ and $P^{\mathrm{j}}=\big\{ p_{k}^{j}|k=1,2,...,n \big\} $ from frames $i$ and $j$, where $n$ is the number of successfully matched feature points. To construct the Gaussian distribution model, we select a certain number of static points, which are considered relatively stable in the scene, and whose optical flow vectors follow a Gaussian distribution.

For each static matching point $(p_{k}^{i},p_{k}^{j})$, we use the Lucas-Kanade optical flow method to calculate its optical flow $V_k$, and then use these optical flow vectors to construct a Gaussian distribution model. The mean $\mu =\frac{1}{n}\sum_{k=1}^n{V_k}$ and the covariance matrix $\Sigma $ of the model are computed based on the optical flow vectors of the static points as follows:

\begin{equation}
  \label{eq9}
	\Sigma =\frac{1}{n-1}\sum_{k=1}^n{(V_k-\mu )(V_k-\mu )^T}\\
\end{equation}

The chi-square value $\chi ^2$ of the vector between two matched points is defined as:
\begin{equation}
  \label{eq10}
	\chi ^2=(V_k-\mu )^T\Sigma ^{-1}(V_k-\mu )
\end{equation}

The chi-squared value has 2 degrees of freedom, and a threshold $ \chi _{0.05}^{2} $ is set. If the calculated chi-squared value falls within this threshold, the point is considered to not deviate significantly from the Gaussian distribution of the static points. Therefore, it can be marked as a static point and mapped back to the Tagged Gaussians, where it will be relabeled as a Static Gaussian.

After completing the labeling of static points and the recovery of Gaussians, we proceed with camera pose optimization using graph-based optimization. The Levenberg-Marquardt (LM) algorithm is employed to minimize the error between the matched Static Gaussians $ P_{i}^{s} $ and feature points $p_i$, where $i\in \mathcal{X} $ represents the set of matched features. This can be expressed as:

\begin{equation}
  \label{eq11}
	\{R,t\}=\mathop {\mathrm{arg}\min} \limits_{R,t}\sum_{i\in \mathcal{X}}{\rho \left( \left\| p_i-\pi _{\left( \cdot \right)}(RP_{i}^{s}+t) \right\| _{\Sigma}^{2} \right)}
\end{equation}
Where $R\in SO\left( 3 \right)$, $t\in \mathrm{R}^3$, $\rho \left( \cdot \right) $  represents a robust kernel function, $\pi \left( \cdot \right) $ is the projection function, and $\Sigma $ is the covariance matrix of the feature point.

\subsection{Dynamic Gaussians Penalty Optimization} \label{sec:3}

\renewcommand{\arraystretch}{1.2}  % 调整表格行高
\begin{table*}[t]
\begin{center}
    \centering
    \caption{\small RESULTS OF METRIC ATE ON TUM DATASET.}
    \caption*{\footnotesize The best-performing results are highlighted in bold, while the second-best results are underscored, “X” denotes the tracking failures. The metric unit is [m]}
    \label{tab1}
    \begin{tabular}{ccccccccccccc}
        \hline
        \multirow{2}{*}{Sequences} & \multicolumn{2}{c}{ORB-SLAM3} & \multicolumn{2}{c}{Photo-SLAM} & \multicolumn{2}{c}{SplaTAM} & \multicolumn{2}{c}{DDN-SLAM} & \multicolumn{2}{c}{NID-SLAM} & \multicolumn{2}{c}{GARAD-SLAM}\\ 
        %\cline{2-13}
        & ATE & STD & ATE & STD & ATE & STD & ATE & STD & ATE & STD & ATE & STD \\
        \hline
        fr3/walking/xyz & 0.3623 & 0.1855 & 0.2862 & 0.1591 & 0.4216 & 0.2538 & \textbf{0.0140} & \underline{0.0085} & 0.0623 & 0.0372 & \underline{0.0155} & \textbf{0.0080} \\
        fr3/walking/xyz\_val & 0.6514 & 0.4354 & 0.6041 & 0.4025 &0.8798 &0.7954 &\underline{0.0132} &\underline{0.0063} & 0.0471 & 0.0146 & \textbf{0.0129} & \textbf{0.0057} \\
        fr3/walking/static & 0.1500 & 0.0850 & 0.1127 & 0.0209 & 0.4612 & 0.2883 & \underline{0.0100} & \underline{0.0052} &  0.0581 &0.0368 & \textbf{0.0061} & \textbf{0.0027} \\
        fr3/walking/static\_val & 0.2856 & 0.0605 & 0.2762 & 0.0615 &0.7418 &0.4571 & \underline{0.0113} & \textbf{0.0076} & 0.0230 & 0.0089 & \textbf{0.0104} & \underline{0.0081} \\
        fr3/walking/rpy & 0.6058 & 0.2781 & 0.5782 & 0.2687 & X & X & \underline{0.0390} & \underline{0.0252} & 0.6080 & 0.4107 &\textbf{0.0360} & \textbf{0.0224} \\
        fr3/walking/rpy\_val & 0.3331 & 0.1769  & 0.3258 & 0.1698 & X & X & \underline{0.0291} & \textbf{0.0176} & X & X & \textbf{0.0280} & \underline{0.0184} \\
        fr3/walking/half & 0.2291 & 0.1013 & 0.2051 & 0.1201 & 0.7321 & 0.6127 & \underline{0.0230} & \underline{0.0132} &  0.6080 & 0.4107 & \textbf{0.0229} & \textbf{0.0124} \\
        fr3/walking/half\_val & 0.3236 & 0.1820 &0.3458 & 0.2048 & 1.1425 & 1.0410 & \underline{0.0243} & \underline{0.0157} & 0.0821 & 0.0477 & \textbf{0.0236} & \textbf{0.0139} \\
        Avg. & 0.3676 & 0.1881 & 0.3418 & 0.1759 & 0.5474 & 0.4310 & \underline{0.0205} & \underline{0.0124} & 0.1861 & 0.1208 & \textbf{0.0194} & \textbf{0.0115} \\

        \hline
    \end{tabular}
    \vspace{-10pt}  % 调整为合适的负值,减少下方间距
\end{center}
\end{table*}

\renewcommand{\arraystretch}{1.2}  % 调整表格行高
\begin{table*}[t]
\begin{center}
    \centering
    \caption{\small RESULTS OF METRIC ATE ON BONN DATASET.}
    \caption*{\footnotesize The best-performing results are highlighted in bold, while the second-best results are underscored, “X” denotes the tracking failures. The metric unit is [m]}
    \label{tab2}
    \begin{tabular}{ccccccccccccc}
        \hline
        \multirow{2}{*}{Sequences} & \multicolumn{2}{c}{ORB-SLAM3} & \multicolumn{2}{c}{Photo-SLAM} & \multicolumn{2}{c}{SplaTAM} & \multicolumn{2}{c}{DDN-SLAM} & \multicolumn{2}{c}{NID-SLAM} & \multicolumn{2}{c}{GARAD-SLAM}\\ 
        %\cline{2-11}
        & ATE & STD & ATE & STD & ATE & STD & ATE & STD & ATE & STD & ATE & STD \\
        \hline
        crowd & 0.4244 & 0.3035 & 0.4568 & 0.3514 & 1.945 & 1.182 & \textbf{0.0180} & \underline{0.0108} & 0.1804 & 0.1021 & \underline{0.0182}	&\textbf{0.0091} \\
        crowd2  & 0.9257 & 0.5681 & 0.9425 & 0.5735 & 3.582 & 1.9664 & \textbf{0.0230} & \textbf{0.0114} & 0.3701 & 0.2559 & \underline{0.0244}	&\underline{0.0123} \\
        person\_tracking & 0.6176 & 0.3291 & 0.5715 & 0.2912 & 0.2598 & 0.2883 & \textbf{0.0430} &\underline{0.0171}& 0.1002 & 0.0612 & \underline{0.0462}	&\textbf{0.0150} \\
        person\_tracking2 & 0.6724 & 0.3254 & 0.6518 & 0.3073 & 0.5117 & 0.4571 & \textbf{0.0380} & \textbf{0.0130} & 0.1473 & 0.0988 &\underline{0.0387}	& \underline{0.0133} \\
        synchronous & 0.7824 & 0.4365 & 0.8071 & 0.4536 &X &X & \underline{0.0398} &\underline{0.0267} & X & X & \textbf{0.0248}	& \textbf{0.0207} \\
        synchronous2 & 1.3425 & 0.4865 & 1.3229 & 0.4695 &X &X & \underline{0.0194} &\underline{0.0128} & X & X & \textbf{0.0069}	& \textbf{0.0038} \\
        balloon & 0.0592 & 0.0418 & 0.0340 &0.0166  &2.4199 &1.0138 & \textbf{0.0180}  &\textbf{0.0094} &0.0355 &0.0186 &\underline{0.0296} & \underline{0.0119} \\
        balloon2  & 0.2030 & 0.1377 & 0.2369 & 0.0735  &1.3569 &0.0724 &0.0410 &0.0267 &\underline{0.0304} &\underline{0.0177} & \textbf{0.0254} &\textbf{0.0118} \\
        Avg. & 0.6284 & 0.3286 & 0.6279 & 0.3171 & 1.2594 & 0.6225 & \underline{0.0300} & \underline{0.0160} & 0.1080 & 0.0693 & \textbf{0.0268} & \textbf{0.0122} \\
        \hline
        % Add more rows here
    \end{tabular}
    \vspace{-5pt}  % 调整为合适的负值,减少下方间距
\end{center}
\end{table*}

\begin{figure*}[htbp]
    \centering
\begin{minipage}[t]{0.05\textwidth}
        \centering
        \raisebox{2cm}[0pt][0pt]{%
            \adjustbox{valign=t}{\rotatebox{90}{\makebox[2cm][c]{\fontsize{7}{9}\selectfont\textbf{ ORB-SLAM3}}}}
        }\\
        \raisebox{0.5cm}[0pt][0pt]{%
            \adjustbox{valign=t}{\rotatebox{90}{\makebox[2cm][c]{\fontsize{7}{9}\selectfont\textbf{ GARAD-SLAM}}}}
        }
    \end{minipage}%
    \begin{minipage}[t]{0.95\textwidth}
        \centering
        \begin{tabular}{cccccc}
        \includegraphics[width=0.14\textwidth]{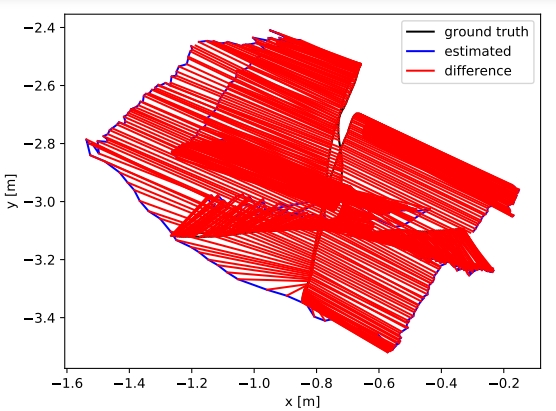}&
        \includegraphics[width=0.14\textwidth]{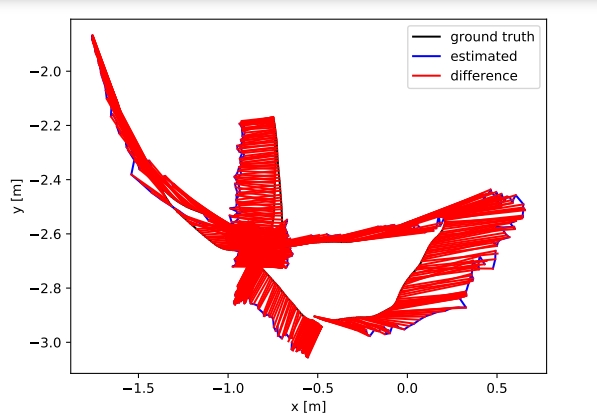}&
        \includegraphics[width=0.14\textwidth]{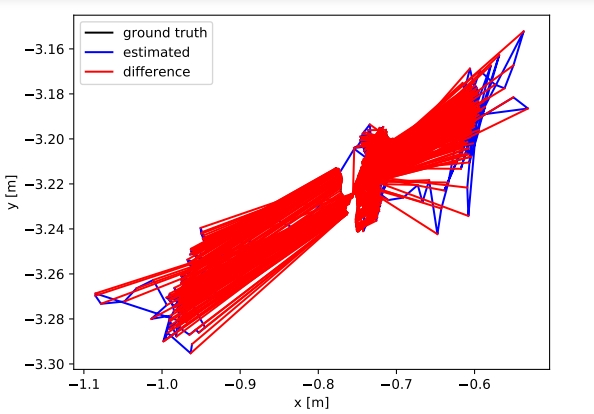}&
        \includegraphics[width=0.14\textwidth]{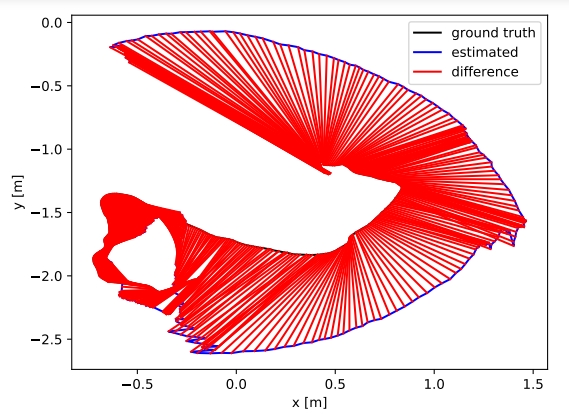}&
        \includegraphics[width=0.14\textwidth]{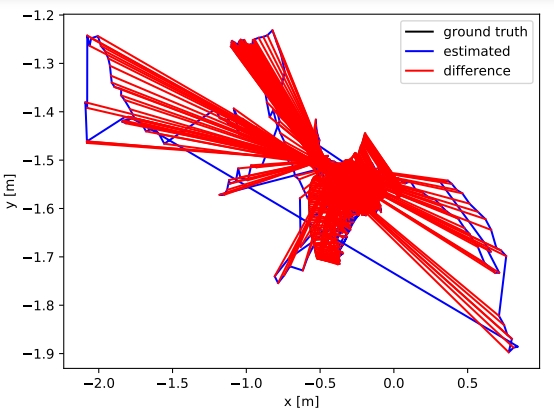}&
        \includegraphics[width=0.14\textwidth]{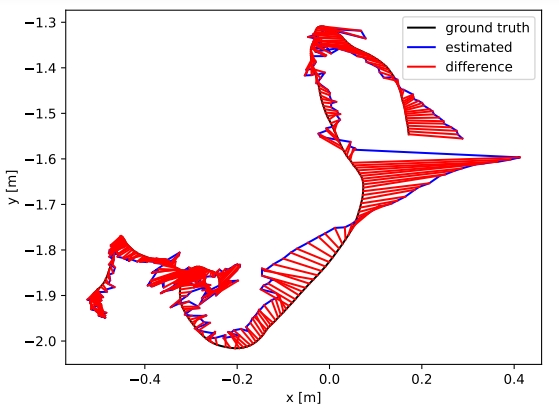}\\
        \includegraphics[width=0.14\textwidth]{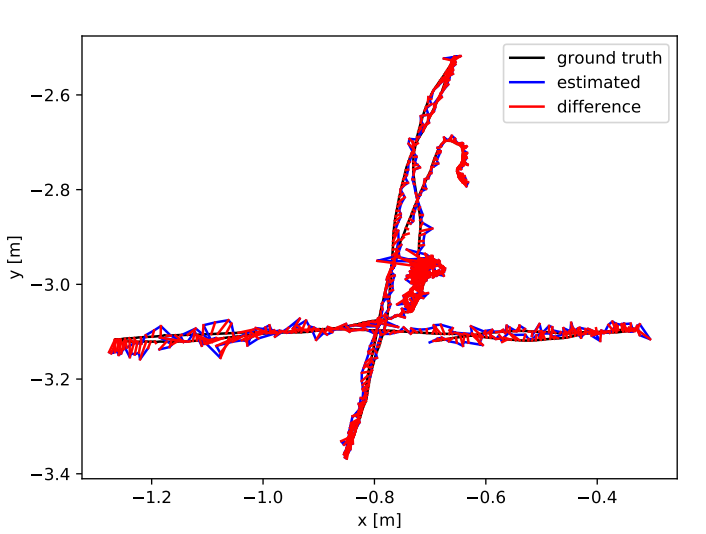}&
        \includegraphics[width=0.14\textwidth]{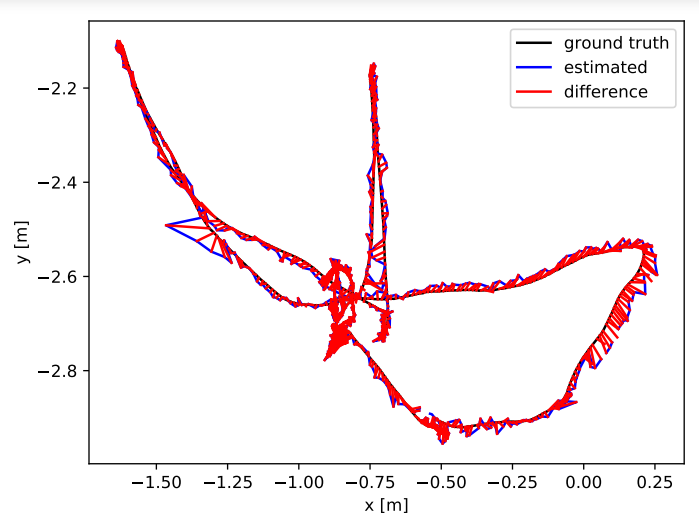}&
        \includegraphics[width=0.14\textwidth]{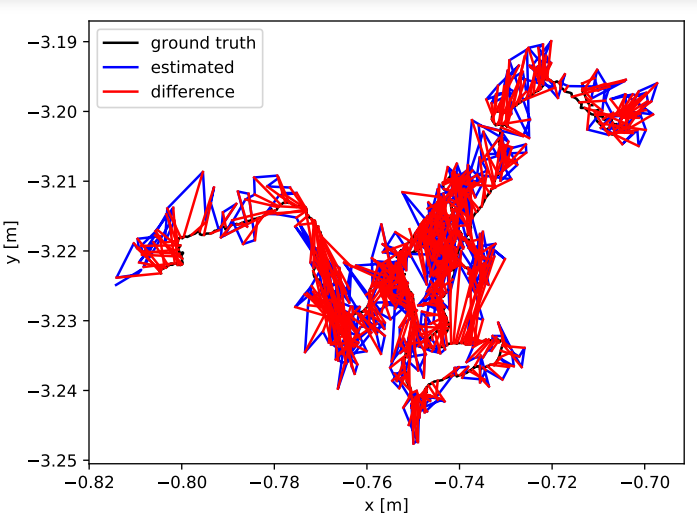}&
        \includegraphics[width=0.14\textwidth]{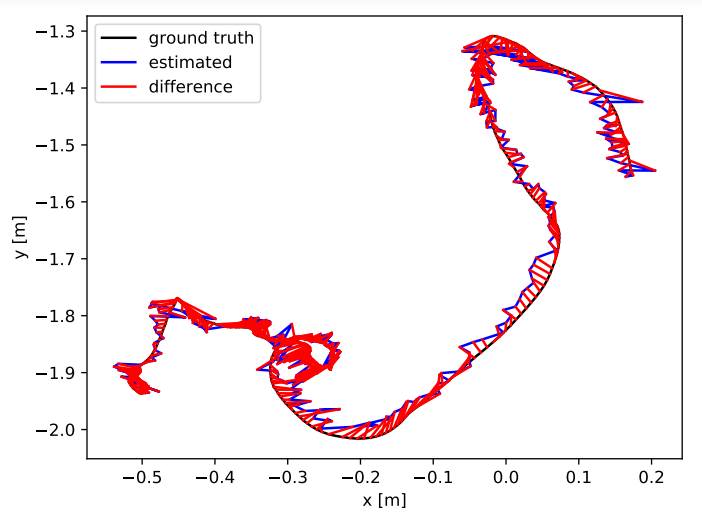}&
        \includegraphics[width=0.14\textwidth]{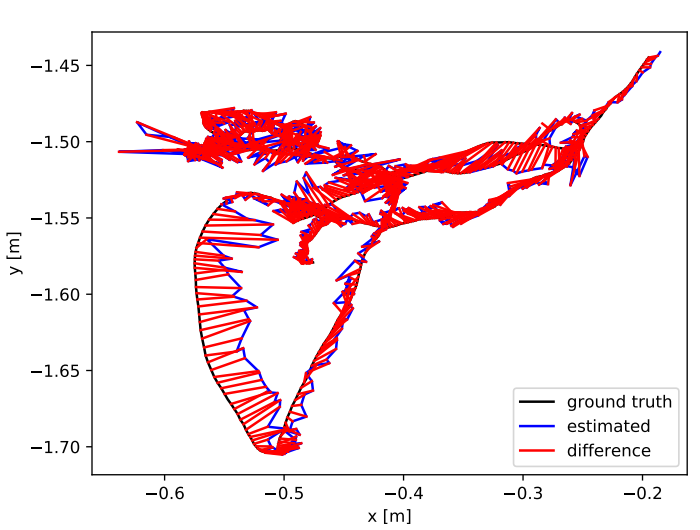}&
        \includegraphics[width=0.14\textwidth]{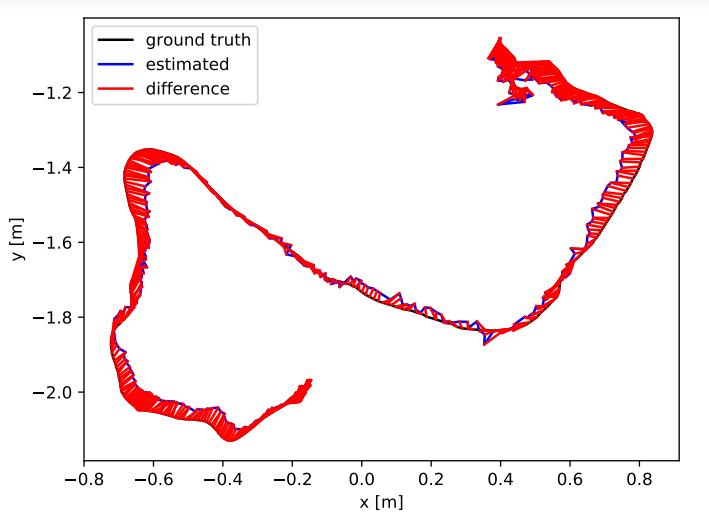}\\
        \fontsize{7}{9}\selectfont \textbf{fr3/walking/xyz} &\fontsize{7}{9}\selectfont \textbf{fr3/walking/half} &\fontsize{7}{9}\selectfont \textbf{fr3/walking/static} &\fontsize{7}{9}\selectfont \textbf{boon\_person\_tracking} &\fontsize{7}{9}\selectfont \textbf{boon\_crowd} &\fontsize{7}{9}\selectfont \textbf{boon\_balloon} \\
        
        \end{tabular}
    \end{minipage}
    \caption{ATE  for ORB-SLAM3 and GARAD-SLAM on some sequences of TUM and BONN datasets.}
    \label{fig3}
    \vspace{-10pt}  % 调整这个数值来减少空白
\end{figure*}

Dynamic Gaussians can significantly affect scene rendering accuracy. The typical approach is to directly remove detected dynamic Gaussians, but this method heavily relies on the precise identification of dynamic Gaussians. If they are not accurately recognized, residual dynamic objects may cause artifacts or blurriness. Conversely, mistakenly removing static Gaussians can lead to the loss of key geometric structures, resulting in holes or incompleteness in the reconstructed scene.

During training with the Gaussian pyramid network, we monitor label changes and prune Gaussians with low opacity and large size to enhance efficiency. However, occlusions or changes in appearance can lead to mistakenly pruning important Gaussians, which is irreversible. To prevent this, we temporarily retain dynamically labeled Gaussians and only delete them after prolonged consistent labeling. Additionally, we introduce a dynamic Gaussians penalty constraint during optimization to minimize the impact of truly dynamic Gaussians on rendering while avoiding the mistaken deletion of static Gaussians, thereby preserving both scene quality and critical information.

The optimization function with a penalty term for dynamic Gaussians not only considers the contribution of dynamic Gaussians to rendering errors but also reduces the interference with the static structure of the scene by balancing the credibility of dynamic labels and the observation duration. The specific optimization function can be expressed as:
\begin{equation}
  \label{eq12}
	\mathcal{L} =\lambda _{p-ssim}\mathcal{L} _{p-ssim}+\lambda _{dyn}\mathcal{L} _{_{dyn}}
\end{equation}
Where $\lambda $ is  the weight coefficient and $\mathcal{L} _{dyn}=\sum_{i\in \mathcal{G} _D}{{\alpha _i}^2} $ .
\vspace{0.7em}  % 增加垂直间距

For $\mathcal{L} _{p-ssim}$, we consider both brightness and SSIM :
% \vspace{-0.3em}  % 减少行间距
\begin{equation}
  \label{eq13}
	\mathcal{L} _{p-ssim}=(1-\lambda )\left| I_r-I_{gt} \right|+\lambda (1-SSIM(I_r-I_{gt}\left)  \right) 
\end{equation}
Where $I_r$ is the rendered image and $T_{gt}$ is the real image.

During the rendering stage, static Gaussians can be rasterized to synthesize corresponding images with keyframe poses. The rendering process is formulated as:

\begin{equation}
  \label{eq15}
    C(R,t)=\sum_{i=1}^N{c_i\alpha _i\prod_{j=1}^{i-1}{(1-\alpha _j)}}	
\end{equation}
Where $N$ is the number of static Gaussians, $c_i$ prensent the color converted from $SH$, and $\alpha _i$ is equal to $\sigma \cdot \mathcal{G} $, $\mathcal{G}$ denotes 3D Gaussian splatting algorithm \cite{13}.

In the Gaussian pyramid-based network optimization process, the Gaussian parameters are first supervised and trained at the top layer of the pyramid, then progressively optimized down through the subsequent layers. This optimization process can be described as:
\vspace{-5pt}  % 调整为合适的负值,减少下方间距
\begin{equation}
  \label{eq16}
  t_i: \mathrm{arg}\min \mathcal{L}(I_r^i, GP^i(I_{gt})), \quad i = n, ...,1, 0
\end{equation}
Where $\mathcal{L} (I_{r}^{i}, GP^i\left( I_{gt} \right) ) $ is the loss function between the rendered image and the ground truth image. $G{P^i}({I_{gt}})$ represents the ground truth image corresponding to the $i-th$ layer of the Gaussian pyramid.

\section{Experiments}

\subsection{Implementation and Experiment Setup}
\textbf{Datasets and Metrics.} We utilize two dynamic datasets, TUM RGB-D \cite{sturm2012benchmark} and BONN RGB-D \cite{palazzolo2019refusion}, to evaluate the performance of our algorithm. The accuracy metric that we assess is the absolute trajectory error (ATE), which represents the global stability of the trajectory. We use root mean square error (RMSE) and standard deviation (STD) to express the robustness and stability of the system. Since it is not feasible to obtain ground truth reconstructions in dynamic environments, we do not compute SSIM or PSNR. Instead, we present a qualitative comparison of the rendering results.

\textbf{Implementation details.} All our datasets were processed on a computer equipped with an Intel Core i9-14900K CPU and an RTX 4080 Ti GPU. The rendering process utilized the stochastic gradient descent (SGD) method, with the Gaussian pyramid set to three levels $(i.e,n=2)$. And the loss weight $\lambda _{p-ssim}=0.8, \lambda _{dyn}=0.2, \lambda =0.2$. We compare our approach with ORB-SLAM3 \cite{campos2021orb}, Photo-SLAM \cite{huang2024photo}, DDN-SLAM \cite{DDN}, NID-SLAM \cite{8}, and SplaTAM \cite{4}, with NID-SLAM being re-implemented by us.

\subsection{Evaluation of Tracking}

Table \ref{tab1} presents the results for eight sequences from the TUM RGB-D dataset. Our approach leverages Gaussian segmentation in dynamic environments, coupled with front-end optical flow verification, resulting in advanced tracking performance compared to existing methods. Table \ref{tab2} illustrates the performance of our system on the more complex BONN RGB-D dataset, demonstrating its superior efficacy in handling challenging scenarios.

Additionally, we present the ATE visualization results for six sequences from the TUM and BONN datasets. As shown in Fig. \ref{fig3}, our system achieves more accurate localization in dynamic environments.

\begin{figure}[htbp]
    \centering
\begin{minipage}[t]{0\textwidth}
        \centering
        \raisebox{5.1cm}[0pt][0pt]{%
            \adjustbox{valign=t}{\rotatebox{90}{\makebox[2cm][c]{\fontsize{5}{7}\selectfont\textbf{ GT}}}}
        }\\
        \raisebox{4cm}[0pt][0pt]{%
            \adjustbox{valign=t}{\rotatebox{90}{\makebox[2cm][c]{\fontsize{5}{7}\selectfont\textbf{ Photo-SLAM}}}}
        }\\
        \raisebox{2.6cm}[0pt][0pt]{%
            \adjustbox{valign=t}{\rotatebox{90}{\makebox[2cm][c]{\fontsize{5}{7}\selectfont\textbf{ SplaTAM}}}}
        }\\
        \raisebox{1.4cm}[0pt][0pt]{%
            \adjustbox{valign=t}{\rotatebox{90}{\makebox[2cm][c]{\fontsize{5}{7}\selectfont\textbf{ NID-SLAM}}}}
        }\\
        \raisebox{0.3cm}[0pt][0pt]{%
            \adjustbox{valign=t}{\rotatebox{90}{\makebox[2cm][c]{\fontsize{5}{7}\selectfont\textbf{ DDN-SLAM}}}}
        }\\
        \raisebox{-1cm}[0pt][0pt]{%
            \adjustbox{valign=t}{\rotatebox{90}{\makebox[2cm][c]{\fontsize{5}{7}\selectfont\textbf{ GARAD-SLAM}}}}
        }
    \end{minipage}%
    \begin{minipage}[t]{0.50\textwidth}
        \centering
        \renewcommand{\tabcolsep}{01pt} % 减小列间距
        \begin{tabular}{cccccc}
        \includegraphics[width=2cm,height=1.5cm]{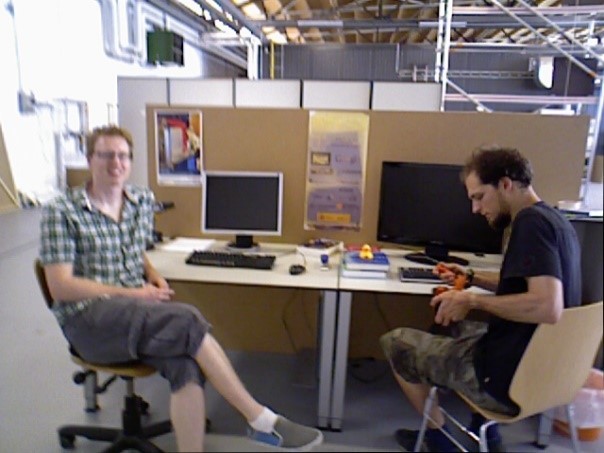}&
        \includegraphics[width=2cm,height=1.5cm]{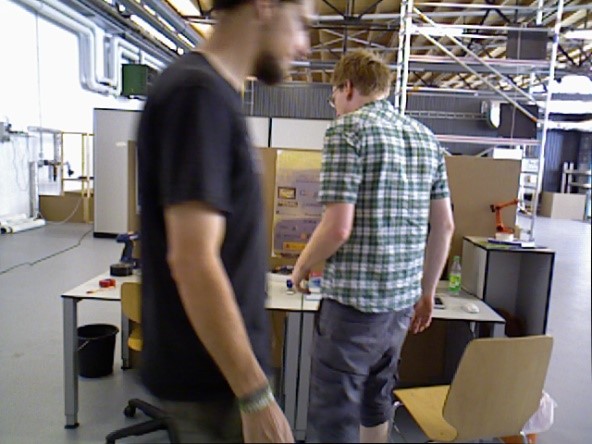}&
        \includegraphics[width=2cm,height=1.5cm]{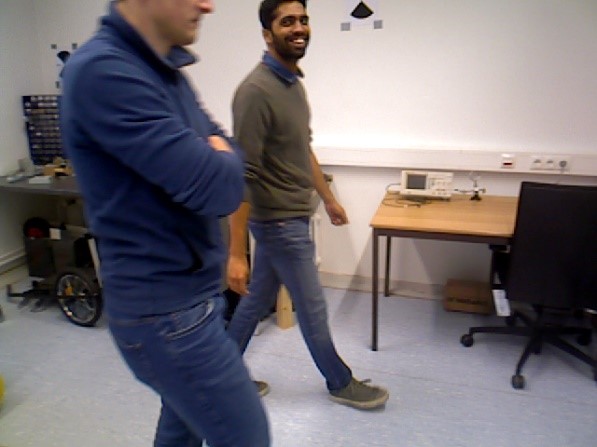}&
        \includegraphics[width=2cm,height=1.5cm]{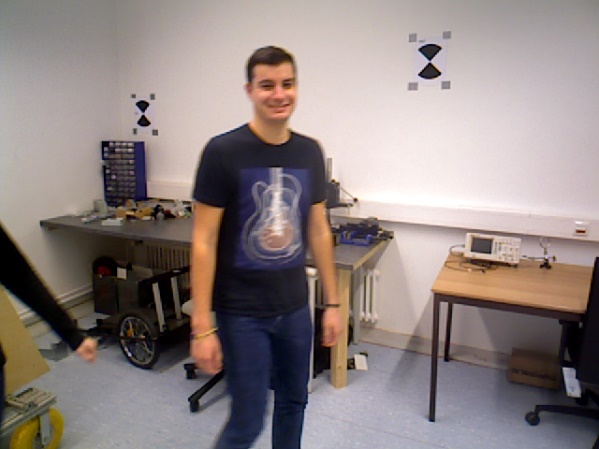}\\
        \includegraphics[width=2cm,height=1.5cm]{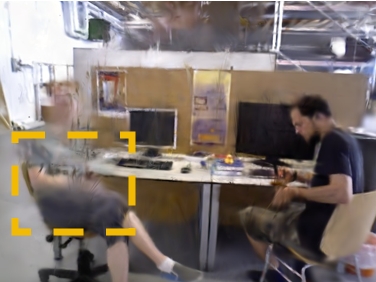}&
        \includegraphics[width=2cm,height=1.5cm]{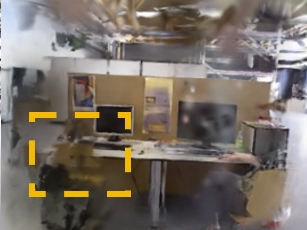}&
        \includegraphics[width=2cm,height=1.5cm]{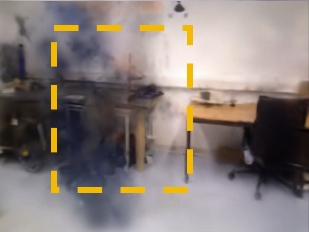}&
        \includegraphics[width=2cm,height=1.5cm]{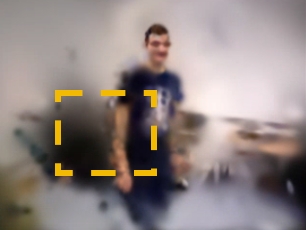}\\
        \includegraphics[width=2cm,height=1.5cm]{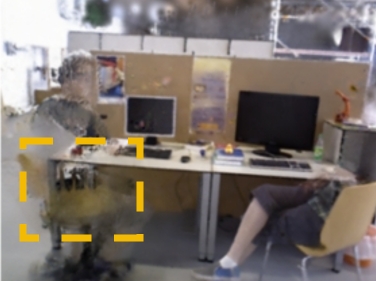}&
        \includegraphics[width=2cm,height=1.5cm]{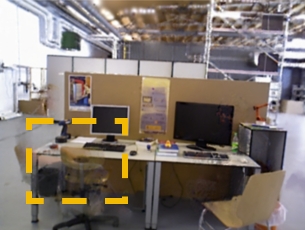}&
        \includegraphics[width=2cm,height=1.5cm]{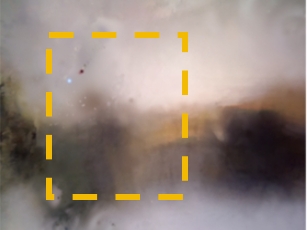}&
        \includegraphics[width=2cm,height=1.5cm]{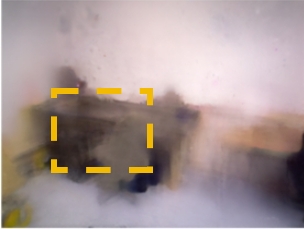}\\
        \includegraphics[width=2cm,height=1.5cm]{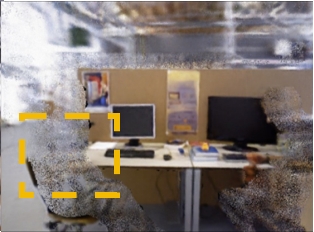}&
        \includegraphics[width=2cm,height=1.5cm]{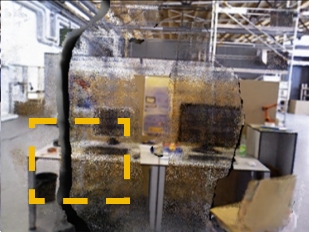}&
        \includegraphics[width=2cm,height=1.5cm]{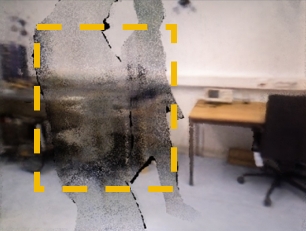}&
        \includegraphics[width=2cm,height=1.5cm]{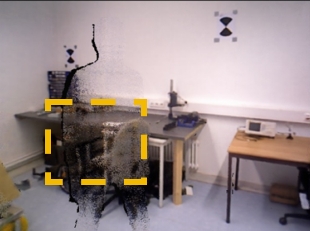}\\
        \includegraphics[width=2cm,height=1.5cm]{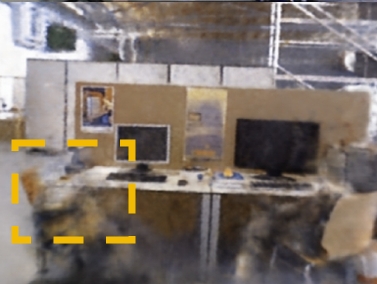}&
        \includegraphics[width=2cm,height=1.5cm]{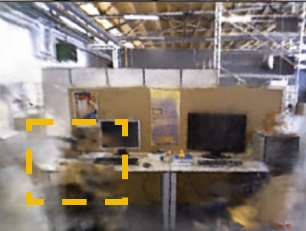}&
        \includegraphics[width=2cm,height=1.5cm]{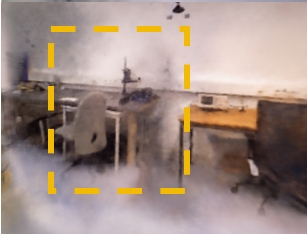}&
        \includegraphics[width=2cm,height=1.5cm]{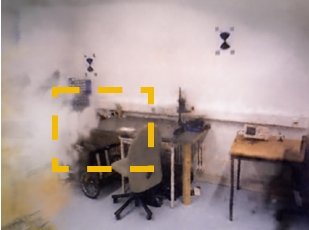}\\
        \includegraphics[width=2cm,height=1.5cm]{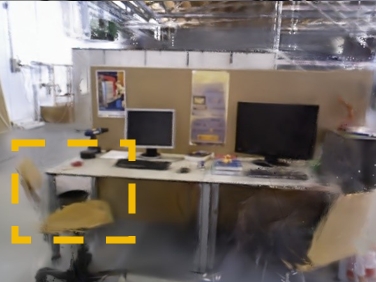}&
        \includegraphics[width=2cm,height=1.5cm]{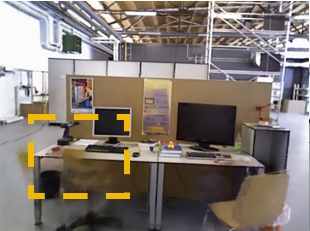}&
        \includegraphics[width=2cm,height=1.5cm]{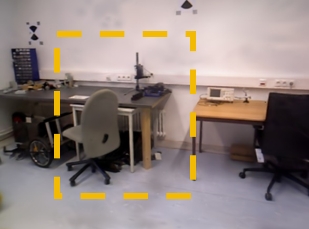}&
        \includegraphics[width=2cm,height=1.5cm]{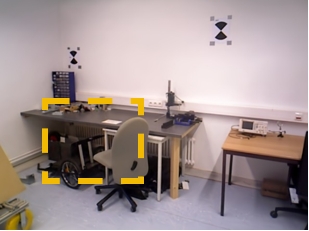}\\
        \fontsize{6}{8}\selectfont \textbf{fr3/walking/xyz\_val}  &\fontsize{6}{8}\selectfont \textbf{fr3/walking/static}  &\fontsize{6}{8}\selectfont \textbf{boon\_crowd} &\fontsize{6}{8}\selectfont \textbf{boon\_syn2} \\
        \end{tabular}
    \end{minipage}
    \caption{Gaussian map of TUM RGB-D and BONN datasets.}
    \vspace{-10pt}  % 调整为合适的负值,减少下方间距
    \label{fig4}
\end{figure}
\subsection{Visualization of Gaussian Map}

To demonstrate the performance of dynamic scene reconstruction, we compared the rendered images with the ground truth poses obtained from the generated Gaussian map, using the same viewpoint to compare with other methods. Four challenging sequences were selected: bonn\_crowd, bonn\_syn2 from the BONN dataset and f3\_walk\_xyz\_val, f3\_walk\_static from the TUM RGB-D dataset. As shown in Fig. \ref{fig1} and Fig. \ref{fig4}, our method demonstrates superior performance in mitigating artifacts in dynamic environments. This highlights the ability of our approach to efficiently segment dynamic Gaussians in real-time, reducing their impact during optimization and achieving improved rendering quality.

\begin{figure}[htbp]
    \centering
    \begin{minipage}[t]{0.16\textwidth}
        \centering
        \includegraphics[width=2.8cm,height=2.1cm]{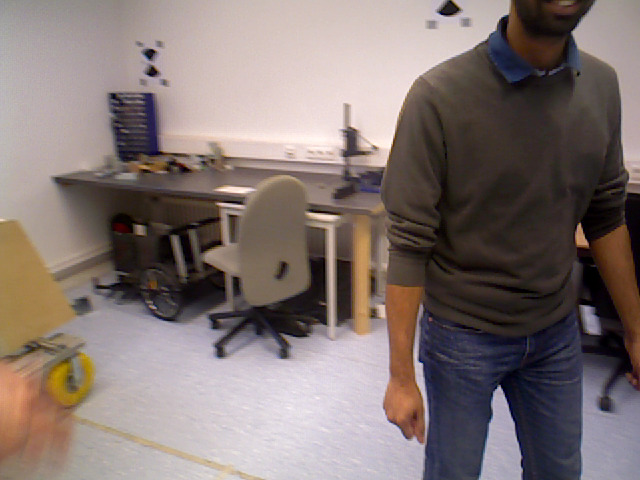} \\
        \fontsize{8}{10}\selectfont \textbf{GT}
    \end{minipage}%
    \begin{minipage}[t]{0.16\textwidth}
        \centering
        \includegraphics[width=2.8cm,height=2.1cm]{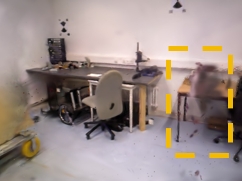} \\
        \fontsize{8}{10}\selectfont \textbf{w/o Mapping}
    \end{minipage}%
    \begin{minipage}[t]{0.16\textwidth}
        \centering
        \includegraphics[width=2.8cm,height=2.1cm]{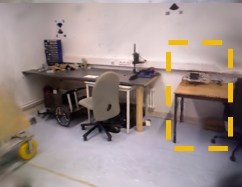} \\
        \fontsize{8}{10}\selectfont \textbf{w/o Full}
    \end{minipage}
    \caption{Comparison of rendering quality.}
    \vspace{-10pt}  % 调整为合适的负值,减少下方间距
    \label{fig5}
\end{figure}

\subsection{Ablation Study}
In this section, we conduct various experiments to verify the effectiveness of our method. w/o CRF refers to the method without CRF segmentation, while w/o Flow indicates performing only CRF segmentation without optical flow verification. w/o Mapping means we do not use the back-end Gaussian mapping strategy but instead employ a traditional approach that only removes front-end feature points. w/o Full refers to performing CRF segmentation along with sparse optical flow verification.

\renewcommand{\arraystretch}{1.2}  % 调整表格行高
\begin{table}[t]
\begin{center}
    \centering
     \caption{\small ABLATION STUDY RESULTS ON TUM AND BONN DATASETS. THE METRIC UNIT IS [m]}
    \label{tab3}
    \begin{tabular}{cccccc}
        \hline
        Dataset & Metric & w/o CRF & w/o Flow  &w/o Mapping  & w/o Full \\
        \hline
        \multirow{2}{*}{TUM} & ATE & 0.3418   & 0.0232  & 0.0212 & \textbf{0.0194} \\
                             & STD & 0.1759  & 0.0154   & 0.0125 & \textbf{0.0114} \\
        \multirow{2}{*}{BONN} & ATE & 0.6287  & 0.0306  & 0.0274 & \textbf{0.0267} \\
                              & STD & 0.3183  & 0.0156  & 0.0127 & \textbf{0.0122} \\
        \hline
    \end{tabular}
    \vspace{-10pt}  % 调整为合适的负值,减少下方间距
\end{center}
\end{table}

\renewcommand{\arraystretch}{1.2}  % 调整表格行高
\begin{table}[t]
\begin{center}
    \centering
    \caption{\small RUNTIME ON FR3/WK/XYZ USING RTX 4080 Ti.}
    \label{tab4}
    \begin{tabular}{cccccc}
        \hline
        &  &NID-SLAM &DDN-SLAM  &SplaTAM  & Ours \\
        \hline
        &Tracking FPS &2.7 &20.5   &52.6  &\textbf{56.4} \\
        &Mapping FPS &3.1 &50.1   &45.5  &\textbf{54.5}  \\
        \hline
    \end{tabular}
    \vspace{-15pt}  % 调整为合适的负值,减少下方间距
\end{center}
\end{table}

We compute the average ATE and STD on the TUM and BONN datasets in Tab. \ref{tab1} and Tab. \ref{tab2} to demonstrate how different methods affect system tracking performance. The results are presented in the Tab. \ref{tab3} indicate that using CRF-based Gaussians segmentation combined with front-end optical flow verification to restore static points facilitates better pose estimation.

For rendering quality, we compared the mapping quality of w/o Mapping and w/o Full. As illustrated in Fig. \ref{fig5}, w/o Mapping, which removes feature points only in the front-end, may lead to artifacts in the back-end rendering. And w/o Full directly applies CRF segmentation on Gaussians in the back-end, effectively reducing the artifacts introduced during the rendering process.
\vspace{-5pt}  % 调整为合适的负值,减少下方间距

\subsection{Time Consumption Analysis}
As shown in Tab. \ref{tab4}, we report the time consumption (frames per second, FPS) of tracking and mapping with other SOTA methods. The results indicate that our method is faster than previous methods.
\vspace{-5pt}  % 调整为合适的负值,减少下方间距

\section{CONCLUSIONS}

We propose GARAD-SLAM, which effectively addresses the issues of tracking drift and mapping errors in 3DGS-based SLAM systems under dynamic scenes. Our approach introduces an anti-dynamic strategy based on back-end Gaussian mapping and front-end sparse optical flow validation. Through step-by-step updates based on neural networks, we achieve accurate removal of dynamic interference and pose correction. We tightly integrate tracking and mapping processes, which mutually enhance each other. Experimental results demonstrate that our method significantly mitigates the impact of transient interference and artifacts compared to other baseline methods, achieving top performance on real-world datasets. Future research will focus on the adaptation of our method for mobile applications and lightweight deployment.

\bibliographystyle{IEEEtran}
\bibliography{IEEEabrv,References}

\end{document}